\documentclass[pmlr,twocolumn,10pt]{jmlr} % W&CP article

\usepackage{booktabs}
\usepackage{siunitx}

\usepackage{inconsolata}
\usepackage{graphicx}
\usepackage{algorithmicx}
\usepackage{algorithm}
\usepackage{algpseudocode}
\usepackage{multirow}
\usepackage{amsmath}
\usepackage{amssymb}
\usepackage{color,soul}

\theorembodyfont{\upshape}
\theoremheaderfont{\scshape}
\theorempostheader{:}
\theoremsep{\newline}

\jmlrvolume{LEAVE UNSET}
\jmlryear{2023}
\jmlrsubmitted{LEAVE UNSET}
\jmlrpublished{LEAVE UNSET}
\jmlrworkshop{Conference on Health, Inference, and Learning (CHIL) 2023} % W&CP title

\title{Token Imbalance Adaptation for Radiology Report Generation}

\author{%
\Name{Yuexin Wu}\Email{ywu10@memphis.edu}\\
\addr University of Memphis
\AND
\Name{I-Chan Huang} \Email{i-chan.huang@stjude.org}\\
\addr  St. Jude Children's Research Hospital
\AND
\Name{Xiaolei Huang}\Email{xiaolei.huang@memphis.edu}\\
\addr University of Memphis
}

\begin{document}

\maketitle

\begin{abstract}
Imbalanced token distributions naturally exist in text documents, leading neural language models to overfit on frequent tokens.
The token imbalance may dampen the robustness of radiology report generators, as complex medical terms appear less frequently but reflect more medical information.
In this study, we demonstrate how current state-of-the-art models fail to generate infrequent tokens on two standard benchmark datasets (IU X-RAY and MIMIC-CXR) of radiology report generation.
% However, no prior study has proposed methods to adapt infrequent tokens for text generators feeding with medical images.
To solve the challenge, we propose the \textbf{T}oken \textbf{Im}balance Adapt\textbf{er} (\textit{TIMER}), aiming to improve generation robustness on infrequent tokens.
The model automatically leverages token imbalance by an unlikelihood loss and dynamically optimizes generation processes to augment infrequent tokens. 
We compare our approach with multiple state-of-the-art methods on the two benchmarks.
Experiments demonstrate the effectiveness of our approach in enhancing model robustness overall and infrequent tokens.
Our ablation analysis shows that our reinforcement learning method has a major effect in adapting token imbalance for radiology report generation.
\end{abstract}

\paragraph*{Data and Code Availability}
In this study, we conduct our experiments in two public datasets, IU X-RAY~\citep{demner2016preparing} and MIMIC-CXR~\citep{johnson2019mimic}. The datasets access to download from \url{https://openi.nlm.nih.gov/} and \url{https://physionet.org/content/mimic-cxr/2.0.0/}. 
We publish our code at
\url{https://github.com/woqingdoua/TIMER}. 
% This initial paragraph is \textbf{mandatory}. Briefly state what data you
% use (including citations if appropriate) and whether the data are
% available to other researchers.\footnote{An example data availability
% statement: This paper uses the MIMIC-III dataset
% \citep{johnson2016mimic}, which is available on the PhysioNet repository.}
% If you are not sharing code, you must explicitly state that you are not
% making your code available. If you are making your code available, then
% at the time of submission for review, please include your code as
% supplemental material or as a code repository link; in either case, your
% code must be anonymized. If your paper is accepted, then you should
% de-anonymize your code for the camera-ready version of the paper. \emph{If
% you do not include this data and code availability statement for your
% paper, or you provide code that is not anonymized at the time of
% submission, then your paper will be desk-rejected.} Your experiments later
% could refer to this initial data and code availability statement if it is
% helpful (e.g., to avoid restating what data you use).

\paragraph*{Institutional Review Board (IRB)}
In this study, we use the publicly available datasets after taking required training courses and signing data usage agreements.
All the publicly available datasets have been de-identified and anonymized.
Our study focuses on computational approaches and does not collect data from human subjects.
We applied the institutional IRB determination that an IRB approval is not required for this study.

\section{Introduction}

\textit{Radiology report generation} is to automatically generate a precise description in natural language given medical images, including computed tomography (CT) and X-RAY images.
Increasing studies have deployed deep encoder-decoder neural architectures to encode medical images and decode the information to generate radiological reports~\citep{jing2018automatic, jing2019show, chen2020generating, chen2021cross, qin2022reinforced}.
Overfitting on frequent tokens is a common challenge in the text generation field that generators fail to predict infrequent tokens~\citep{yu2022rare}. 
Our empirical analysis has demonstrated that over 80\% medical terms are infrequent tokens, while frequent tokens can count over 82\% corpus (Section \ref{sec:data}). 
The complex and lengthy tokens naturally occur less frequently than simple words~\citep{nikkarinen2021modeling}, which is common across medical scenarios~\citep{demner2016preparing, johnson2019mimic}.
However, existing studies have not explicitly consider infrequent tokens for radiology report generation, which can decrease model robustness and lead to imprecise reports.

%there are studies discussing the imbalance, but no prior studies have proposed method on the token imbalance in the medical report generation
Methods to reduce token-frequency-overfit commonly deploy post-processing~\citep{mu2018allbutthetop} or regularization techniques~\citep{sean2019neural, wang2020Improving, yu2022rare} for token embeddings.
However, the approaches usually work for text-to-text generation given text sentences as inputs, while medical images are inputs in our study.
\citeauthor{nishino2020reinforcement} proposes reinforcement learning approach for the class imbalance issue. 
However, the approach aims to solve image category imbalance instead of token imbalance.
Modeling the token imbalance for the multimodal generation task is an unsolved challenge, especially for medical scenarios.

% our method
In this study, we propose the Token Imbalance Adapter
(TIMER) model using reinforcement learning~\citep{sutton1999policy} to adapt infrequent token generation and evaluate on radiology report generation by two publicly available datasets, IU X-RAY~\citep{demner2016preparing} and MIMIC-CXR~\citep{johnson2019mimic}.
Our approach deploys an unlikelihood loss to penalize incorrect predictions for frequent tokens and develops a dynamic adaptation module to adjust the optimization process automatically.
We compare the TIMER with three state-of-the-art baselines on overall performance and show overall improvements of our approach. 
By evaluating the performance of low and high-frequent tokens, our approach can significantly improve generation performance on  infrequent token sets and maintain stable performance on frequent tokens.
% summarize what experiments you have done
% We demonstrate the effectiveness of our proposed method on two publicly available datasets.

Our major contributions are summarized as follows: 1) to our best knowledge, this is the first study that explicitly adapts token imbalance for radiology report generation. We have demonstrated the importance of infrequent tokens in radiology datasets, as most medical terms occur infrequently (Figure \ref{fig:medical_token}). 2) We propose a reinforcement learning method that effectively improves model overall performance as well as performance of infrequent tokens. 3) We conduct extensive ablation analysis to illustrate the effectiveness of adapting infrequent tokens. The ablation analysis proves that our method successfully reduces generation biases on frequent tokens and dynamically leverage token frequency.

\section{Data}
\label{sec:data}

We retrieved two publicly available datasets of radiology report generation, IU X-RAY~\citep{demner2016preparing} and MIMIC-CXR~\citep{johnson2019mimic}. 
1) \textit{IU X-RAY}, collected by Indiana University, provides 3,955 reports and 7,470 X-RAY images from Indiana Network for Patient Care.
2) \textit{MIMIC-CXR} (denote as MIMIC) is the largest radiography and publically available dataset to date. The dataset contains 227,835 reports with 377,110 images collecting between 2011 and 2016 at the Beth Israel Deaconess Medical Center.
Each radiology report may associate with one or more front and side X-rays images. 
We summarize the data statistics in Table~\ref{tab:data}.

\begin{table}[htp]
\centering
\caption{Statistics of the two radiography datasets. Length is the report's average length, and Vocab is the vocabulary size.}
\resizebox{0.48\textwidth}{!}{
\begin{tabular}{c|c|c|c|c}
& Images & Reports & Length & Vocab \\\hline\hline
IU X-RAY &7,470 &3,955 &35.99 &1,517\\
MIMIC  &377,110 &227,835 &59.70 &13,876\\
\end{tabular}}
\label{tab:data}
\end{table}

% discuss how the infrequent tokens can be important as well.
Overfitting on frequent patterns is a common challenge for deep neural models ---  radiology report generators can easily overfit on frequent tokens than infrequent tokens.
A recent text generation study~\citep{yu2022rare} has found that text generators commonly perform much worse on infrequent tokens.
The low performance on infrequent tokens can significantly impact radiological report generation, where many important medical terms do not appear frequently.
In this study, we argue that \underline{\textit{infrequent tokens matter}} in radiology report generation.

\begin{figure}[htp]
\centering 
\caption{Ratios of medical terms across five equal splits of vocabulary. 1 represents the most infrequent token set, and 5 refers to the most frequent token split.}
\includegraphics[width=0.482\textwidth]{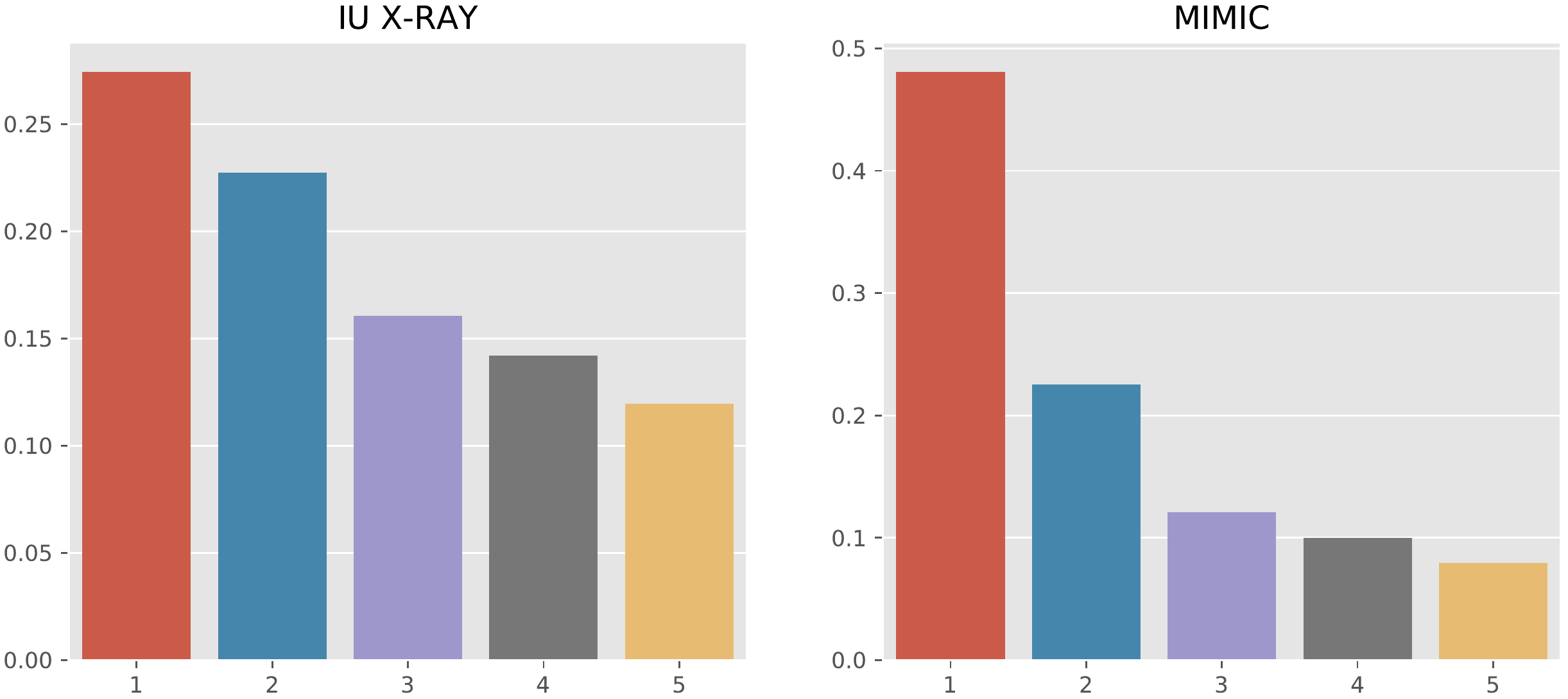}
\label{fig:medical_token}
\end{figure}

To verify our claim, we conduct a quantitative analysis of medical terms across token distributions.
First, we deploy a named entity recognition (NER) model (en\_ner\_bc5cdr\_md) from \texttt{Sci-SpaCy} to extract disease-related terms.\footnote{Package details refer to the link: \url{https://allenai.github.io/scispacy/}. We choose the NER model because it achieves the best performance on the medical data and covers larger medical terms than the other available models.}
Next, we split the vocabulary of each dataset into five equal segments.
Our empirical counts show that the first 20\% (segment) of vocabulary types account for over 82\% tokens in each dataset while the rest of vocabulary types (80\%) only account for less 18\% of corpus tokens.
We calculate frequency ratios of medical terms in each segment and visualize the results in Figure~\ref{fig:medical_token}.
The figure shows that infrequent tokens contain more medical terms than frequent tokens, especially for complex tokens (average character length per token).\footnote{In IU X-RAY, the top medical terms in frequent tokens are pleural, effusion, heart, lung, and focal, while the top ranks in the infrequent tokens are diverticula, shrapnel, hemorrhage, prosthesis, arthroplasty.}

Ignoring the low performance on infrequent tokens can significantly harm robustness of radiological generators (i.e., baselines fail in infrequent token evaluations in Table~\ref{tab:im}).
% While the frequent tokens have higher ratios of medical terms, the infrequent tokens still contain a large portion of medical terms: majority of medical terms fall into the infrequent tokens.
This is essentially important for medical scenarios where infrequent tokens are more likely as complex domain terms, which are not commonly as frequent words~\citep{nikkarinen2021modeling}.
However, there is few study in the text generation task adapting infrequent tokens, especially for the radiological report generation.
Thus, the issue inspires us to propose our model, \textbf{T}oken \textbf{Im}balance Adapt\textbf{er} (\textit{TIMER}).

\begin{figure*}[ht!]
\caption{Illustration of our proposed TIMER model. We use arrows to indicate model workflow. Blue arrows refer to loss value calculations by Equations~\ref{treward} and \ref{uloss}. Parameter update processes follow red dotted lines. Our learning progress has inner and outer loops. In the inner loop, we update the NLG model by $\mathcal{L}_{NLG}+\mathcal{L}_{UL}$. In the outer loop, we implement the dynamic adaptation (DA) via reinforcement learning and update DA module ($\eta$) by the updated parameters ($\theta_{new}$) and the reward $r_{t}$ in Eq.~\ref{total_r}.}
\includegraphics[width=1\textwidth]{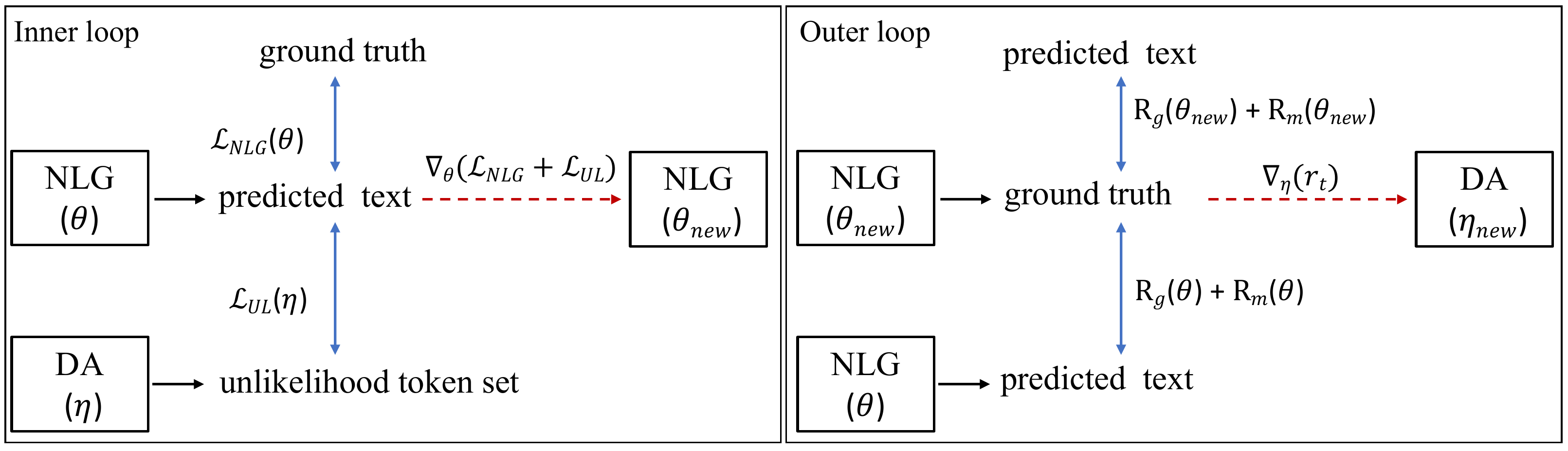}
\label{mr}
% \vspace{-0.15in}
\end{figure*}

\section{Token Imbalance Adapter}

In this section, we present our approach \textit{TIMER} in Figure~\ref{mr}.
TIMER consists of three major modules, 1) unlikelihood loss, 2) dynamic adaptation, and 3) joint optimization.
%The report generator is a generation pipeline that inputs radiology images and generates text reports.
We deploy unlikelihood loss to reduce overfit and token imbalance of the text generator by penalizing a frequent-token set.
The dynamic adaptation module deploys reinforcement learning to allow adjust the frequent-token set automatically.
Finally, We elaborate on how to jointly optimize the text generator and dynamic adaptation module.

\subsection{Problem Statement}
Radiology report generation is an image-to-text generation task.
Given a radiology image $\bf x$ and  the corresponding report $\boldsymbol{y}=\left(y_{0}, \ldots, y_{L}\right)$ with length $L$, the task aims to train a text generator ($p_{\theta}(\mathbf{y} \mid \bf x)$) by minimizing
\begin{equation}\label{treward}
    \mathcal{L}_{NLG}(\theta) = - \sum_{l=1}^{L} \log p\left(y_{l} \mid y_{1}, \ldots, y_{l-1}, \mathbf{x} ; \theta\right)
\end{equation}, 
where $p\left(y_{l} \mid y_{1}, \ldots, y_{l-1}, \mathbf{x} ; \theta\right)$ is from the softmax function for the next token prediction, and $\theta$ refers to generation model parameters.
Token imbalance can cause prediction overfit that leads to low performance on low-frequent tokens.
To solve the challenge, we balance the performance by the unlikelihood loss.

%In the test stage, the generation is a recursive progress, 
%\begin{equation}
%    p(\mathbf{y} \mid \mathbf{x})=\prod_{t=1}^{T} p\left(y_{t} \mid y_{1}, \ldots, y_{t-1}, \mathbf{x}\right).
%\end{equation}

\subsection{Unlikelihood Loss}\label{fixed_set}
%Inspired by the work~\citep{sean2019neural}, we propose an Unlikelihood Token Module (DA). The module designs a sequence-irrelevant unlikelihood loss to control biased predictions due to the token imbalance distribution. The unlikelihood loss aims to reduce overfitting effects by penalizing predicted probabilities for frequent tokens across generation process. Given a set of frequent tokens $\mathcal{U}_h$ as an unlikelihood token set in a corpus,  we define the unlikelihood loss function as:
Inspired by the work~\citep{sean2019neural}, we unitize unlikelihood loss to reduce over-fitting effects by penalizing predicted probabilities for frequent tokens. Firstly, we calculate the average predicted possibility $p(u)$ for each token  $u$ in a report,
\begin{equation}\label{avg_prob}
    p(u) = \frac{\sum_{l=1}^{L} \log \left(p\left(u \mid  y_{1}, ..., y_{l}\right)\right)}{L}
\end{equation}
Then, given a set of frequent tokens $\mathcal{U}_h$ in a corpus, the unlikelihood loss punishes each token $u \in \mathcal{U}_h$ by
\begin{equation}\label{uloss}
    \mathcal{L}_{\mathrm{UL}}( \mathcal{U}_h)=-\sum_{u \in \mathcal{U}_h} \log (1-p\left(u)\right)
\end{equation}
To decrease $\mathcal{L}_{\mathrm{UL}}$, the model predicts lower probabilities $p(u)$ on the frequent tokens, $u$. 
However, the unlikelihood loss has two issues: frequent token set and combining optimization objectives. Because statically fixing the frequent token set and equally combining two optimization objectives ($\mathcal{L}_{NLG}$ and $\mathcal{L}_{\mathrm{UL}}$) are not ideal for the report generation task.
For example, Section~\ref{subsec:dynamic} shows a static set of frequent tokens is not effective to reduce the token-frequency overfit of report generators. To enable dynamic adaptation, we deploy reinforcement learning to select a frequent token set.
We leverage different training objectives by the joint optimization (Section~\ref{subsec:joint}).

\subsection{Dynamic Adaptation}
We deploy a reinforcement learning (RL)~\citep{sutton1999policy} method allowing for a dynamic unlikelihood token set instead of a fixed set.
The dynamic adaptation includes three major components, a policy network, a value network, and a reward.
The dynamic adaptation paradigm is to adjust the unlikelihood token set and reduce token frequency effects during the report generation.
Specifically, the NLG model is jointly optimized by the generation and unlikelihood losses, where the DA module decides the unlikelihood set. Then our imbalance reward can evaluate the improvements between the trained and untrained model with the unlikelihood loss. This reward helps the DA module to dynamically tune a better unlikelihood set, which can guide the NLG model to better training by balancing performance between infrequent and frequent tokens.
Such a setting increases flexibility and dynamics of adjusting balances between frequent and infrequent tokens.

\paragraph{Policy Network,} $\pi_{\theta}\left(a_{t} \mid s_{t}\right)$, aims to predict a frequent token set as an action $a_{t}$ from a state $s_{t}$ at step $t$. We take a dot product of token distributions between prediction in Eq.~\ref{avg_prob} and ground truth in a sample as our state.
The dot product can reflect total token-level prediction errors.
The policy network feeds a state ($s_{t}$) into the fully connected layer and sigmoid function and outputs a possibility estimation for an action $a_{t}$.  
The fully connected layer has the same input and output sizes as the vocabulary size. 
Finally, the policy network adopts a Bernoulli sampling on the estimated possibility vector and obtains the final unlikelihood token set $\mathcal{U}$. 
The policy network identifies the frequent token set for each sample through learning stages. 
This setting can help the generation model dynamically adjust the tokens' weight to reduce overfitting during optimization. 
% \hl{We describe the policy network optimization in section}~\ref{subsec:joint}.

\paragraph{Value Network,} A2C algorithm~\citep{sutton1999policy}utilizes a value network to guide the policy network's 
 learning and reduce variance due to the environment's randomness. In our case, the NLG model may overfit to different tokens after each optimization step, which causes the same policy can have a different reward. To overcome such randomness, we also introduce a value network to help the policy network's  stable learning.
 $Q(s_{t},a_{t})$, is to predict a reward by given a state ($s_{t}$) and a action ($a_{t}$). In our case, $Q(s,a)$ is a 1-D convolutional network with two input channels and one output channel. The kernel size was set to 10 in our experiment. The predicted reward is the average of the convolutional network output.
 % \hl{We describe the value network optimization in section}~\ref{subsec:joint}.

\paragraph{Reward}\label{para:reward} We design a reward to combine generation report quality and leverage the performance variations between frequent and infrequent tokens.
Our reward consists of a text generation reward ($R_{g}$) from Eq.\ref{treward} and and imbalanced evaluation reward ($R_{m}$). 
The imbalanced evaluation divides the comparison count of performance variations into four sets according to the tokens' frequency and calculates the F1 score for each token's set, respectively.
We denote $F(\mathcal{U}_{l})$, $F(\mathcal{U}_{m})$ and $F(\mathcal{U}_{h})$ as the F1 score of the low, medium and high-frequency token sets, respectively. 
%For example, we denote $\mathcal{U}_{l}$ as low frequency token set and the corresponding F1 score (denoted $F(\mathcal{U}_{l})$) can be obtained as the following,
%\begin{equation}\label{TP}
%\begin{aligned}
%    TP &= \{y_{t} \in \mathcal{U}_{l}\text{, for each } y_{t} \in \mathbf{y}_{prediction}\} \\
%    &\cap \{y_{t} \in \mathcal{U}_{l}\text{, for each } y_{t} \in \mathbf{y}_{truth}\} \\
%\end{aligned}
%\end{equation}
%\begin{equation}
%FP = \{y_{t} \in \mathcal{U}_{l}\text{, for each } y_{t} \in \mathbf{y}_{\text{prediction}}\} - TP
%\end{equation}
%\begin{equation}
%FN = \{y_{t} \in \mathcal{U}_{l}\text{, for each } y_{t}\} -  TP
%\end{equation}
%\begin{equation}
%\text{precision} = \frac{|TP|}{|TP|+|FP|}
%\end{equation}
%\begin{equation}
%\text{recall} = \frac{|TP|}{|TP|+|FN|}
%\end{equation}
%\begin{equation}
%F(\mathcal{U}_{l}) = \frac{2*\text{precision}*\text{recall}}{\text{precision}+\text{recall}}
%\end{equation}
To avoid F1-score overestimation due to token repetitions, we restrict each token in the prediction only project into one token in the ground truth instead of the token repetition. For example, if the prediction text includes repetitive toke such as ``bone is is intact'' and the ground truth is ``the heart size is abnormal'', the correct prediction token number is 1 in this case according to our definition. Because the correct prediction token (``is'') only occurs one time in the ground truth.
%Similarly, we obtain medium and high frequency token sets F1 score, $F(\mathcal{U}_{m})$ and $F(\mathcal{U}_{h})$ respectively.
Then, we average the F1 score of  $F(\mathcal{U}_{l})$, $F(\mathcal{U}_{m})$, and $F(\mathcal{U}_{h})$ as our imbalanced reward $R_m$:
% \begin{equation*}
%     R_{m} = \frac{F(\mathcal{U}_{l})+F(\mathcal{U}_{d})+F(\mathcal{U}_{h})}{3}
% \end{equation*}
\begin{equation*}
    \begin{split}
    R_{m} = (|F(\mathcal{U}_{h}) - F(\mathcal{U}_{l})| + |F(\mathcal{U}_{m}) - F(\mathcal{U}_{l})| \\ + |F(\mathcal{U}_{h}) - F(\mathcal{U}_{m})| ) / 3
    \end{split}
\end{equation*}
Compared to calculating the F1 score in one token set, this method can alleviate a biased evaluation due to imbalanced token distribution since $R_{m}$ can balance performance across different frequency token sets.  
% The reward is the sum of $R_{g}$ and $R_{m}$.
% To better compare, we add a reward from the last step model NLG($\theta$) as our baseline.
We can formulate the final reward as follows:
\begin{align}
\label{total_r}
    r_{t} = &R_{g} (\theta_{new}) + R_{m} (\theta_{new}) - R_{g} (\theta) - R_{m} (\theta) 
\end{align}
where $R_{g}$ is text generation loss in Eq.\ref{treward}, and $\theta$ refers to model parameters. 
We include optimization steps in the following section to learn the policy network and value network with the reward.

\subsection{Joint optimization}
\label{subsec:joint}
Our optimization includes \textit{inner} and \textit{outer} loops. In the inner loop, we update the natural language generation (NLG) model with parameters $\theta$ according to the loss $\mathcal{L}_{NLG}$ in Eq.~\ref{treward}) and the unlikelihood loss $\mathcal{L}_{\mathrm{UL}}$ in Eq.~\ref{uloss} as follows,
\begin{equation}\label{inner_loss}
    \mathcal{L}_{inner} = \mathcal{L}_{NLG}(\theta) + \mathcal{L}_{\mathrm{UL}}(\mathcal{U};\eta)
\end{equation}
%where $\mathcal{U}$ is the unlikelihood toke set, and $\eta$ indicates DA parameters. We can optimize and update the NLG model parameters ($\theta$) by,
\begin{equation}\label{inner_loss2}
    \theta_{new}= \theta-\nabla_{\theta} \mathcal{L}_{inner}
\end{equation}

In the outer loop, we update the dynamic adaptation module (DA) with parameters $\eta$ by A2C algorithm~\citep{sutton1999policy}.
The policy network learns how to interact with the environment within time $t$ by minimizing the following expectation:

% Then we calculate loss by updated parameter $\theta_{new}$ to update DA parameters based on reinforcement learning theory.
%TIMER utilizes the dynamic adaptation module to decide whether a token should be included in the unlikelihood token set, which allows us to dynamically adjust the set size.
%Our model optimization deploys A2C~\citep{sutton1999policy} to handle the dynamic adjustment, where the policy function $\pi_{\theta}$ learns how to interact with the environment within time $t$ by minimizing the following expectation:
\begin{equation*}
     \mathcal{L}_{policy} = - ( \mathbb{E}[\sum_{t}  \log \pi_{\theta}\left(a_{t} \mid s_{t}\right) A(s_{t}, a_{t})]),
\end{equation*}
where $A(s_{t}, a_{t})$ is an advantage estimate and equals to the real advantage in expectation. 
A2C utilizes Temporal-Difference(TD) to calculate advantage $A(s_{t}, a_{t})$,
\begin{equation*}
    A(s_{t}, a_{t}) = r_{t}+\gamma Q\left(s_{t+1}, a_{t+1}\right)-Q\left(s_{t}, a_{t}\right),
\end{equation*}
where $r_{t}$ is a reward after taking the action $a_{t}$ in the state $s_{t}$, and $Q\left(s_{t}, a_{t}\right)$ is a value function. 
% We show the reward calculation in section~\ref{para:reward}.
%The action aims to predict whether a token is in unlikelihood set in a sample and $a_{t} \in  R^{B \times 1}$, where $B$ is vocabulary size.
$ \gamma$ is a discount factor that denotes the trade-off between immediate rewards and future returns. 
We predict an action for each sample. 
Each sample has an individual unlikelihood token set, therefore each step action is independent. Thus, we do not need to consider a future return and the $A(s_{t}, a_{t})$ calculation can be simplified as follows,
\begin{equation*}
     A(s_{t}, a_{t}) = r_{t}-Q\left(s_{t}, a_{t}\right)
\end{equation*}
%We take the dot product of token distribution between prediction in Eq.~\ref{avg_prob} and ground truth in a sample as our state.
Next, we optimize value function $Q(s_{t}, a_{t})$. 
A2C trains a value network by minimizing MSE loss of TD,
\begin{equation*}
 \mathcal{L}_{value} = (r_{t}-Q\left(s_{t}, a_{t}\right))^{2}
\end{equation*}
%%%%%%%%%%%%%%%%%%%%%%Explain two questions here:

% 1. why the loss function calculation is not computationally effective?

% 2. how equation 5) transit to the 7) is not discussed.

%%%%%%%%%%%%%%%%%%%%%%End
Then, we add an entropy loss as regularization term to promote action diversity of the policy function, 
\begin{equation*}
    \mathcal{L}_{entropy}=-\sum \mathrm{P}(\pi_{\theta}) \log  \mathrm{P}(\pi_{\theta}).
\end{equation*}
We integrate the multiple losses together as the outer loss and update DA module as follows:
\begin{equation}
\label{outer_loss}
% \begin{aligned}
    \mathcal{L}_{outer} = \mathcal{L}_{policy} + \mathcal{L}_{value} + \mathcal{L}_{entropy}
% \end{aligned}
\end{equation}
DA module parameter is optimized by,
\begin{equation}
\label{outer_loss2}
    \eta_{new}=\eta-\nabla_{\eta} \mathcal{L}_{\text {outer }}\left(\theta_{new}\right)
\end{equation}
We show the detailed optimization process of in \ref{alg:Framwork}.

\begin{algorithm}[ht!] 
\caption{Optimization Process of TIMER.}
\label{alg:Framwork} 
\begin{algorithmic}[1]
\Require
	% The training set $\mathbf{x}_{s}$, maximum iteration $I$, the iteration of inner loop $N$;
	% \For{$i=1$; $i<I$; $i++$}
	% \For{$n=1$; $n<N$; $n++$}
        The training set $\mathbf{x}_{s}$, maximum iteration $I$, the iteration of inner loop $N$;
            for $i=1$; $i<I$; $i++$;
	    for $n=1$; $n<N$; $n++$
    \State Samples a batch from $\mathbf{x}_{s}$;
    \State Update $\theta$ via Eq.~\ref{inner_loss} and Eq.~\ref{inner_loss2} ;
    % \EndFor
    \State Samples a batch from $\mathbf{x}_{s}$;
    \State Update $\eta$ via Eq.~\ref{outer_loss} and Eq.~\ref{outer_loss2};
    % \EndFor
\end{algorithmic}
\end{algorithm}

%\paragraph{DA architecture} DA includes two networks, a policy network and the network $A(a,s)$. 
% add a sentence of rm model architecture function The policy network is a fully connected layer with the same input and output size as the vocabulary size ($B$). The policy network feeds an average possibility prediction $p(u)$ for each token in a sentence from the NLG model into the fully connected layer and the sigmoid function. Then we can obtain a possibility estimation for an action $a_{t}$.  We adopt a Bernoulli sampling on the estimated possibility vector and obtain the final unlikelihood token set $\mathcal{U}$. The value network $A(s,a)$ is a 1-D convolutional network, where has two input channel and one output channel. The kernel size set to 10 in our experiment. The predicted reward is the average of the convolutional network output.

\begin{table*}[htp]
\centering 
\caption{Performance summary. $\Delta$ indicates averaged percentage improvements of TIMER over baselines. Clinical Metric calculates the F1 score.} \label{overallp}
\begin{tabular}{c||ccccccc}

Methods & BLEU\_1 &BLEU\_2 & BLEU\_3 & BLEU\_4 & Meteor& Rouge\_L & Clinical Metric \\\hline\hline
\multicolumn{7}{c}{IU X-RAY} \\ \hline
BiLSTM & 41.83 & 29.30 & 21.27 & 15.49 & 18.75 & 34.26 &65.06 \\
R2GEN  &48.80 &31.93 &23.24 &17.72 &20.21 &37.10 &63.62\\
CMN &45.53  &29.50 &21.47 &16.53 &18.99 &36.78 &64.83\\
CMM+RL &49.30 & 30.08 & 21.45 & 16.10 & 20.10 & 38.20 &40.79\\
TIMER & \bf49.34 & \bf32.49 & \bf 23.84 & \bf 18.61 &\bf 20.38 &\bf 38.25 &\bf94.40\\\hline
$\Delta$ &6.42	&7.57	&9.07	&13.06	&4.45	&4.55	&61.16 \\
\hline\hline
\multicolumn{7}{c}{MIMIC} \\ \hline
BiLSTM &26.81 &15.77 &10.12 &7.00 &11.26 &26.00 &49.50\\
R2GEN&35.42 &21.99 &14.50 &10.30 &13.75 & 27.24 &34.77\\
CMN &35.60 &21.41 &14.07 &9.91 &14.18 &27.14 &41.21\\
CMM+RL &38.10 &22.10 &14.45 &10.02 &14.53 &27.66 &28.36\\
TIMER & \bf38.30 &\bf 22.49 &\bf 14.60 &\bf 10.40 &\bf14.70 &\bf 28.00 &\bf75.86\\
 \hline
$\Delta$ &11.27	&9.66	&9.01	&10.50	&8.64	&3.54	&49.30\\
\end{tabular}
\end{table*}

\section{Experiment}
% data details

We follow the previous studies~\citep{jing2018automatic, chen2020generating, chen2021cross} for data preprocessing, data splits (training, development, and test splits), and model evaluations.
We use natural language generation (NLG) metric and clinical efficacy to evaluate our model and baseline, such as BLEU \citep{papineni2002bleu}, METEOR \citep{denkowski2011meteor}, and ROUGE-L~\citep{lin2004rouge}.
% We use CheXpert to label our generation report and ground truth and then compare the labels to evaluate clinical efficacy (CE) metrics by F1 score in the MIMIC dataset.
To evaluate clinical efficacy, we utilize CheXpert~\citep{irvin2019chexpert} to annotate generated reports for MIMIC and IU X-RAY.
While there are other evaluation methods (e.g., RadGraph~\citep{jain2021radgraph} and CheXbert~\citep{smit2020combining}), we deploy the same metrics as the baselines for consistence.
More details of data preprocessing and our implementations are in the Appendix, which allows for experiment replications.
%we fine-tuned a pre-trained medical classification model BioBert-PubMed200kRCT~\citep{deka2022evidence} with the labels from Medical Subject Heading (MESH)\footnote{\url{nlm.nih.gov/mesh/meshhome.html}}.
% We set the learning rate as 1e-4 and train 50 epochs. 
% The accuracy is 98.65\% in the test set.
% Then we compare the labels of our generation report and ground truth predicted by our fine-tuned model.

\subsection{Baselines}
To demonstrate the effectiveness of our model, we compare our approach with four state-of-the-art (SOTA) baselines\footnote{We chose the SOTA methods that achieved the best performance during our experimental steps. \textbf{Note} that this direction evolves rapidly in recent years, and therefore there might be newer methods published during our study's submission and review.} that use the same experimental settings, BiLSTM~\citep{jing2018automatic}, R2Gen~\citep{chen2020generating}, CMN~\citep{chen2021cross}, and CMM+RL~\citep{qin2022reinforced}.
To ensure comparisons, we utilized their open-sourced models and followed their experimental settings.

\textbf{BiLSTM} \citep{jing2018automatic} incorporates semantic tags into visual feature representations to generate radiology reports. To obtan the semantic tags, BiLSTM utilizes the Convolutional Neural Networks (CNNs) to predict semantic tags and corresponding visual abnormality regions, which later will be fed to the Long Short-Term Memory (LSTM) for report generation. 
The model utilizes a co-attention mechanism to localize regions containing abnormalities and generate narrations.
% We use the released code to train our own models on the two datasets.
% We implement this baseline by inputting images impression and finding texts as training samples and evaluate it on finding texts in IU X-Xray. 
% Since MIMIC does not have conclusive impressions, we train this model only on long report texts.

\textbf{R2Gen} \citep{chen2020generating} designs a memory-driven Transformer, where a relational memory is used to capture critical information in the generation process. 
Then the model proposes a memory-driven conditional layer normalization to incorporate the memory into the decoder of the Transformer. 
We reuse the code and released models from the study.

\textbf{CMN} \citep{chen2021cross} proposes a shared memory mechanism to
record the alignment between images and texts so as to facilitate the interaction and generation across modalities. 
The memory mechanism is an intermediate medium and contains querying and mapping processes that enhance and smooth mappings between text and image representations.
We use the author's released code and model to reproduce the results.

\textbf{CMM + RL} \citep{qin2022reinforced}  proposes a method for enhancing text and image representation alignments using reinforcement learning (RL). 
The RL treats the generation model as the agent that interacts with an external environment (image and text representations).
The method provides appropriate supervision from NLG evaluation metrics to search for better mappings between features from different modalities.
Our TIMER model uses the RL in a different way that our RL is to dynamically adapt token imbalance instead of aligning modalities.
% Upon generating the end-of-sequence (EOS) token, reward the generation model based on evaluation metrics.

\begin{table}[htp]
\centering
\caption{The imbalanced evaluation in the high- and low-frequency token set. We evaluate the F1 score by dividing tokens into high and low-frequency set with three different bucket size, (e.g., 1/8 represents top 1/8 frequent tokens as a high-frequency set).}
\label{tab:im}
\resizebox{0.48\textwidth}{!}{
    \begin{tabular}{clcccccc} \hline
    && \multicolumn{2}{c}{IU X-RAY} & \multicolumn{2}{c}{MIMIC} \\
     Ratio & Method & low & high & low  & high  \\\hline\hline
    \multirow{4}{*}{1/8}&Bi-LSTM  & 3.85 & 47.31 & 1.27 & 37.48 \\ 
    &R2GEN &4.46 &\bf 62.73 &2.52 &52.01 \\
    &CMN  & 5.88 & 55.86 & 2.23 & 45.60 \\
    &CMM + RL &5.19	&49.36 & 0.21 &43.64\\
    &TIMER &\bf 13.23 & 61.89 &\bf 3.15 &\bf 52.66 \\\hline
    
    \multirow{4}{*}{1/6}&Bi-LSTM  &1.97&44.06 &0.89 &30.77 \\
    &R2GEN &2.80 &61.62 &\bf2.00 &49.86\\
    &CMN  &5.75 &65.12	&0.85 &\bf 52.02 \\
    &CMM + RL &5.08	&49.26 & 0.14 &43.36\\
    &TIMER  &\bf5.93 &\bf 67.79 &\bf 2.02 &51.72   \\\hline
    
    \multirow{4}{*}{1/4}&Bi-LSTM  &0.00 &44.41 &0.28	&30.09\\ 
    &R2GEN &1.16 &59.98  &0.00  &48.77\\
    &CMN  &2.60 &63.92 &0.33 &51.09 \\
    &CMM + RL &2.19	&47.21 & 0.07	&43.05\\
    &TIMER   &\bf 8.66 &\bf 64.00 &\bf 0.58 &\bf 51.39  \\
    
    \hline\hline
    %Delta-BiLSTM $(\%)$ & 166.11 & 17.68 & 148.77 & 40.50 \\
    %Delta-R2GEN $(\%)$ & 196.37 & -1.34 & 23.24 & 3.81 \\
    %Delta-CMN $(\%)$ & 125.05 & 10.78 & 41.23 & 15.49
    \end{tabular}
}
\end{table}

\section{Results and Analysis}
This section provides an overview of the performance by both natural language generation (NLG) metrics and clinical efficacy~\citep{irvin2019chexpert}. We also present an imbalanced evaluation focusing on high and low frequencies token sets. Furthermore, we perform an ablation analysis to assess the impact of individual modules in our approach.

\subsection{Overall Performance}
Table~\ref{overallp} presents overall performance and the details of improvements percentage are shown in Table~\ref{tab:delta_all} in the Appendix.
The results show that our approach significantly outperforms baselines, ranging from 3.54\% to 61.16\% improvements. 
% Notably, since our experiment employs the same preprocessing method as R2Gen and CMN, we report the results directly from their original publications.
% Our model consistently outperforms the baselines in terms of the $BLEU_1$ metric, with only marginal changes in other metrics.
% This outcome is attributed to our proposed RL module, which enables the model to improve uni-gram prediction accuracy for both common and rare tokens. 
% Our approach maintains similar performance levels for other metrics while delivering significant improvements for infrequent token generations (Section~\ref{IP}).
Compared to the BiLSTM, the transformer-based models (R2GEN, CMN, CMM$+$RL) consistently perform well on all datasets among the four baseline generators. However, while CMM$+$RL's performance is on par with other baselines (e.g.,). Its BLEU\_4 scores are relatively lower, indicating inefficiency in generating fluent sentences. One reason could be that CMM$+$RL employs a greedy sampling approach in self-critic learning, which fails to consider long-term returns. In contrast, our model significantly improves language fluency by achieving a 13.06\% and 10.50\% increase in BLEU\_4 scores on IU-XRAY and MIMIC, respectively. 
Our model's most significant improvement is in the clinical metric, which we attribute to its focus on improving the prediction of infrequent tokens, particularly clinical vocabulary.

\subsection{Imbalanced Performance} \label{IP}

% Our model addresses the issue of biased token generation, which often results from imbalanced token distributions. 
Table~\ref{tab:im} presents a performance evaluation on infrequent tokens, which demonstrates effectiveness of our approach to learning token imbalance. 
We use F1-score to evaluate model performance of high and low-frequency token sets. 
We define three high-frequency by three levels, $1/4$, $1/6$, and $1/8$, which refers to the percentage of top frequent tokens as high-frequency in the vocabulary and the rest are low frequent tokens.
The setting is to demonstrate effectiveness of our approach in adapting imbalanced tokens.

By comparing the baselines, we find that our method significantly outperforms them in the low-frequency token set, both on IU X-RAY and MIMIC datasets by 1\% to 245.6\%. 
Our model improves performance of rare tokens by over 100\% on IU X-RAY dataset and 41\% on MIMIC dataset, highlighting the importance of addressing token imbalance issue. Neural models trained on skewed token distributions tend to overfit on frequent tokens, resulting in a greater prediction error for rare tokens.
Furthermore, our method does not sacrifice model performance on the frequent token set. Compared to CMN, our approach achieves a 10.78\% improvement on IU X-RAY and a 15.49\% improvement on MIMIC with a ratio of $1/8$.
%Our model has a higher improvement on the MIMIC than the IU X-ray. We infer this as the MIMIC has more training entries than the IU X-ray.

\begin{table}[ht]
    \centering
    \caption{Ablation analysis. ``-DA+UL'' denotes removing DA and using unlikelihood loss with a fixed token set. ``-DA'' denotes the model is trained by negative log-likelihood loss~\ref{treward} without unlikelihood loss. F1 is the clinical metric.}\label{ab}
    \begin{tabular}{cccc}
    \hline
        \multicolumn{4}{c}{IU X-RAY}  \\ \hline
        ~ & -DA + UL & -DA  & full \\ \hline
        BLEU\_1 & 44.73 & 47.91 & 49.34 \\ 
        BLEU\_2 & 29.08 & 30.60 & 32.49 \\ 
        BLEU\_3 & 21.08 & 21.76 & 23.84 \\ 
        BLEU\_4 & 16.36 & 15.98 & 18.61 \\ 
        METEOR & 18.72 & 19.73 & 20.38 \\ 
        ROUGE\_L & 35.41 & 36.41 & 38.25 \\
        F1 &82.43 &89.72  &94.40 \\\hline
        \multicolumn{4}{c}{MIMIC}  \\ \hline
        ~& -DA + UL & -DA  & full \\ \hline
        BLEU\_1 & 34.02 & 35.31 &38.30 \\ 
        BLEU\_2 & 20.90 & 21.02 & 22.49 \\ 
        BLEU\_3 & 13.99 & 13.73 & 14.60 \\ 
        BLEU\_4 & 10.02 & 9.54 & 10.40 \\ 
        METEOR & 13.76 & 14.03 & 14.69 \\ 
        ROUGE\_L & 27.43 & 26.66 & 28.00 \\
        F1 &68.86 &70.28 &75.86\\\hline
    \end{tabular}
\end{table}

\subsection{Ablation Analysis}
We performed an ablation analysis to evaluate the effectiveness of individual modules in our model. We used the notation ``-DA+UL'' to denote removing the DA module and replacing the module with a fixed frequent token set for the unlikelihood loss. 
We empirically selected the top 100 frequent tokens in IU X-RAY and the top 600 frequent tokens in MIMIC as the token set size. 
We used the same evaluation metrics as in the previous sections and summarized the performance results in Table~\ref{ab}.

Our results show that both the unlikelihood loss and DA modules can improve model performance across all metrics, proving the effectiveness of the proposed modules in generating radiology reports. However, we observed that adding the DA module has more performance improvements compared to the unlikelihood loss. This suggests that dynamic adaptation has a greater contribution in promoting model robustness overall and infrequent token adaptation.

\subsection{Dynamic Adaptation Analysis}
\label{subsec:dynamic}

\begin{figure}[htp]
\caption{The performance comparison with different sizes of unlikelihood token sets. F1 is the clinical metric.}\label{top_token}
\centering 
\includegraphics[width=0.482\textwidth]{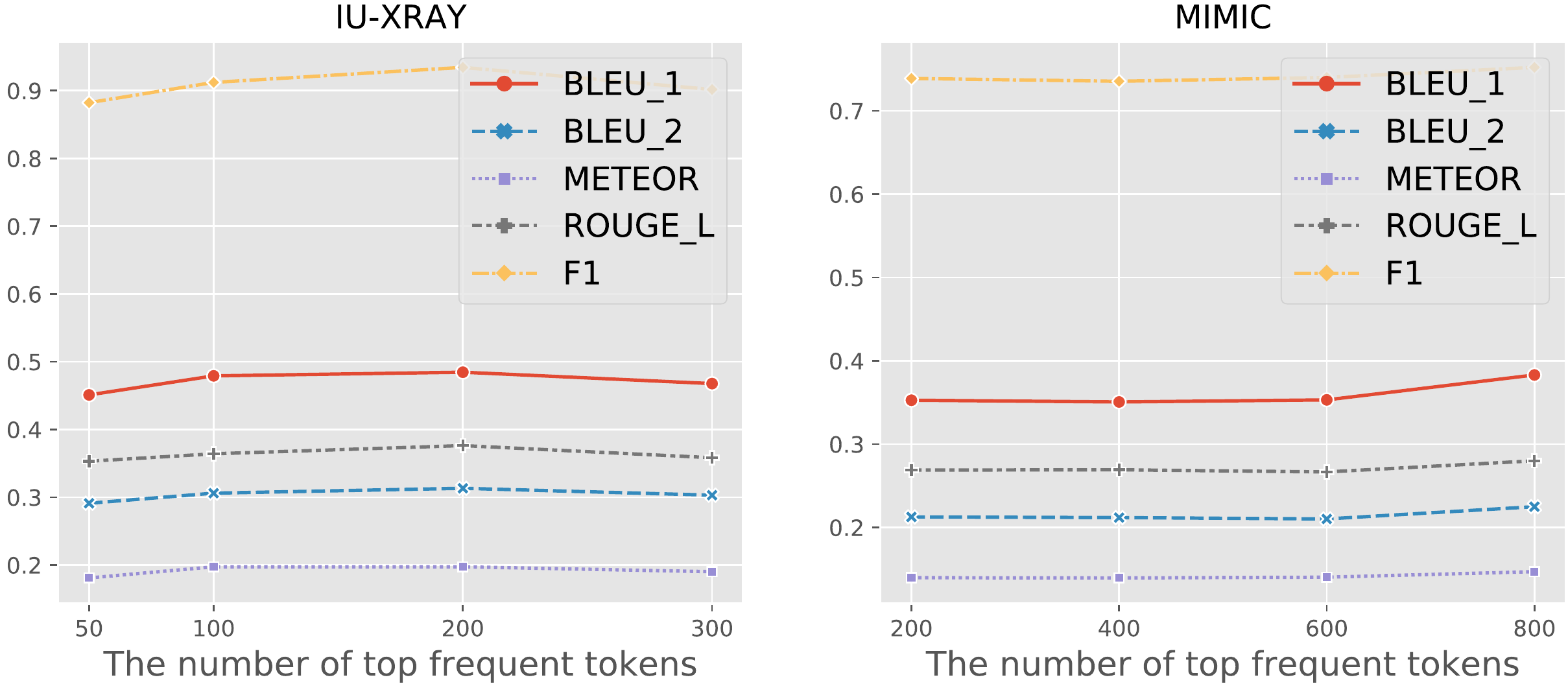}
\end{figure}

TIMER dynamically adjusts the size of the unlikelihood token set to improve generator performance. 
This ablation analysis is to evaluate dynamic adaptation by comparing dynamic and static adjustments.
For the static adaptation, the size of the unlikelihood token set is fixed at different sizes for the IU X-RAY and MIMIC datasets. 
The results of changes in the fixed sizes are visualized in Figure~\ref{top_token}. 
The results show that different fixed sizes yield lower BLEU, METEOR, and ROUGE scores than dynamic adaptation, indicating that the size of the unlikelihood token set impacts generator performance. 
The effectiveness of dynamic adaptation provides further evidence of TIMER can improve overall performance and incorporate imbalance token adaptation.

\begin{figure*}[ht!]
\caption{Qualitative comparison between TIMER and CMN. We highlight correct predictions of infrequent tokens. We set top 20\% tokens in the vocabulary as frequent and the rest as infrequent tokens.}
\includegraphics[width=1\textwidth]{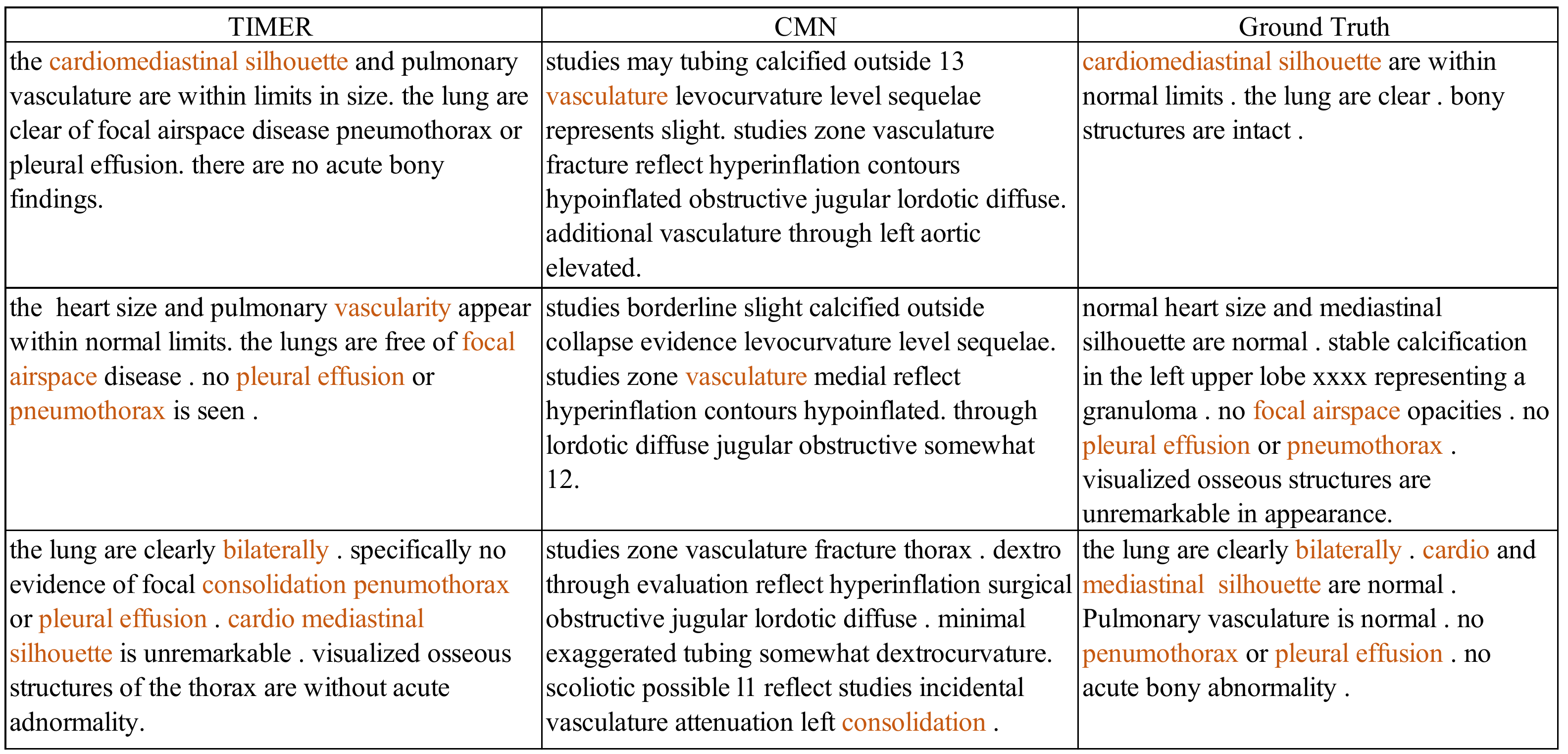}
\label{fig:qs}
% \vspace{-0.15in}
\end{figure*}

\subsection{Qualitative Analysis}
To further investigate the effectiveness of our model's generation, we performed a qualitative analysis by selecting three generated samples from IU X-RAY and MIMIC-CXR datasets and compared them with the ground truth, as shown in Figure~\ref{fig:qs}. Our analysis revealed that TIMER can generate descriptions that closely match the ground truth. Furthermore, compared to CMN, TIMER generated more medical terms such as ``pleural effusion,'' ``pneumothorax,'' and ``mediastinal silhouette.'' Additionally, TIMER accurately diagnosed the medical condition and produced a semantically similar sentence to the ground truth. For instance, the sentence ``there are no acute bony findings" generated by TIMER was semantically similar to the ground truth sentence ``bony structures are intact.''

\section{Related Work}

\paragraph{Radiology report generation} is to generate descriptive text from radiology images. 
An encoder-decoder network is the primary neural architecture of the task. 
For instance, \citep{jing2018automatic} built a multi-task learning framework that employs a hierarchical LSTM model to generate long radiology reports, and a co-attention mechanism to jointly perform the prediction of tags. 
Recent studies deployed Transformer architecture~\citep{vaswani2017attention} to improve the task performance~\citep{lovelace2020learning, chen2020generating, chen2021cross}. 
% \citet{lovelace2020learning} used the Transformer to generate text descriptions and extract clinical observations from generated reports to reach an agreement between the prediction and ground truth.
For example, \citep{chen2020generating} proposed a relational memory to record key information in the generation process and applied a memory-driven conditional layer normalization to incorporate the memory into the decoder of the Transformer. 
\citep{chen2021cross} designed a shared memory between the encoder and decoder to record the alignment between images and texts. 
Several recent works~\citep{dalla2022multimodel,mli2022cross, yang2022knowledge} have enhanced the quality of text generation by developing models that can learn clinical knowledge directly from reports. For instance, \cite{yang2022knowledge} devised an automatic information extraction mechanism to extract clinical entities and relations directly from training reports, while Dalla et al.\citep{dalla2022multimodel} extracted entities and relations from images and then generated complete reports based on the extracted entities and relations.
To better evaluate the factualness of generated reports, \citep{smit2020combining} trained a BERT-based model to label the reports' diseases. 
More recently, Some works~\citep{miura2021improving,delbrouck2022improving,qin2022reinforced} have incorporated reinforcement learning (RL) into radiology reports to improve their quality. Delbrouck et al.\citep{delbrouck2022improving} improved the quality of report generation by predicting more precise entities, while Qin et al.~\citep{qin2022reinforced} improved report generation performance by better aligning the images and text with RL. 
%However, none of these works have considered the issue of overfitting to frequent tokens. Therefore, in this work, we present a novel reward that can aid the NLG model in overfitting to frequent tokens, thereby enhancing the quality of radiology report generation.
However, none of these studies have explicitly worked on token imbalance, which is the main focus of our study.

\paragraph{Reinforcement Learning (RL)} in NLG tasks is to improve sequence prediction~\citep{shi2018toward, bahdanau2016actor, hao2022tetacher}.
The actor-critic approach proposed by ~\citeauthor{bahdanau2016actor} considers the task objective during training, improving maximum likelihood training but suffering from sparse reward.
To address this, \citeauthor{shi2018toward} uses the maximum entropy Inverse Reinforcement Learning to mitigate the issues of reward sparsity.
\cite{wu2022unsupervised} employs reinforcement learning to address label imbalance issues across various domains. However, this is a classification task and it is challenging to apply it directly to generation tasks.
In this study, we employ an actor-critic approach to optimize our model, but instead of applying it to the NLG model directly, we propose a novel learning strategy that updates the unlikelihood token set dynamically.
While \citep{nishino2020reinforcement} applies RL to radiology report generation, a key difference is that they focus on document label imbalance rather than the token imbalance, which is the primary target of our study.
% As far as we know, our study is the first to investigate reinforcement learning for token imbalance in radiology report generation.

\paragraph{Imbalance modeling} refers to the task of modeling skewed distributions. Various strategies have been proposed to handle imbalanced data in natural language processing tasks, such as oversampling, under-sampling, and the few-shot technique~\citep{tian2021embedding, yang2020hscnn}. 
While existing solutions focus on classification tasks, those techniques may not apply to radiology report generation, a multimodal image-to-text generation in healthcare. 
In radiology report generation, the imbalance between normal and abnormal samples has been addressed in prior studies through methods such as data augmentation with reinforcement learning~\citep{nishino2020reinforcement} and using separate LSTMs for abnormal and normal sentence generation~\citep{harzig2019addressing}. However, no previous study has focused explicitly on imbalanced token distributions in medical report generation. In this work, we propose a novel approach that leverages reinforcement learning techniques to address this issue.

\section{Conclusion}

In this study, we have proved the importance of infrequent tokens in radiology report generation and proposed a reinforcement-learning method to adapt token imbalance. We demonstrate effectiveness of our approach (TIMER) over three state-of-art baselines on radiology report generation. 
TIMER automatically penalizes overfitting on frequent tokens and dynamically adjusts rewards on infrequent token generations. Extensive experiments and ablation analysis show that TIMER can obtain significant improvements on infrequent token generations while maintaining performance on frequent tokens by multiple evaluation metrics.
While we evaluate our approach on radiology report generation, we expect its broad applicability to text generation tasks and will extend applications in our future work.
% Our code repository will be available at [URL].

% \newpage

\section*{Limitations}
While we have proved the effectiveness of our proposed method, three primary limitations must be acknowledged to appropriately interpret our evaluation results, task and preprocessing.

% task limitation.
\paragraph{Task.} We primarily focus on the task of radiology report generation, while the task is only one of the downstream evaluations in the text generation field.
Experiments on the radiological benchmarks may not generalize to all text generation tasks, and infrequent tokens may not contain critical and complex medical terms.
In this study, to ensure consistency, we compare with the SOTA baselines on the same datasets of radiology report generation.
\textbf{Note} that this task has evolved rapidly in recent years, and therefore there might be newer methods published during our study's publishing processes.
% We will add newer studies if their proceedings have been fully released.
% For example, the official EMNLP-2022 proceeding was only available in Feb. 2023 while the conference was held in Dec. 2022.
% Future directions may expand our study over other text generation tasks and datasets, such as style transfer~\citep{jin2022deep}.

\paragraph{Preprocessing.} We utilize the datasets and follow the same preprocessing steps from the previous study~\citep{chen2020generating}.
Existing studies~\citep{nguyen2021automated, qin2022reinforced} may have variations in the preprocessing steps that can directly impact the final results.
For example, \citep{nguyen2021automated} selected the top 900 frequent tokens for performance evaluations, which shows a significant improvement in the frequent tokens.
To ensure a consistent comparison, we keep the same experimental settings and deploy the released models from the baselines~\citep{jing2018automatic, chen2020generating, chen2021cross} that reported state-of-the-art performance on the same datasets.
% The processing steps in the baselines may also exclude some rare tokens, which occur less than a threshold.
Our future work will develop more comprehensive evaluations of rare tokens across the existing models.

\paragraph{Human Evaluation.} It is necessary to invite radiologists for human evaluations. 
However, we did not include the approach due to \textit{subjectivity} and \textit{domain} challenges.
A common approach is to sample limited generated reports from a pair of methods~\citep{miura2021improving, kurisinkel2021coherent}. 
However, the same baselines with different human evaluators may yield varied evaluation results~\citep{liu2021contrastive, liu2021competence}.
It is a challenge to have enough certified radiologists to evaluate the task.
We conducted preliminary human evaluations by the qualitative analysis in Table~\ref{fig:qs}, though we miss enough support from radiologists. 
Having human evaluations will be our future work to provide more comprehensive perspectives.

\section*{Ethic and Privacy Concerns}

We follow data agreement and training procedures to access the two radiology report datasets.
To protect user privacy, we have followed corresponding data agreements to ensure proper data usage and experimented with the de-identified data. 
Our experiments do not store any data and only use available multimodal entries for research demonstrations.
Due to privacy and ethical considerations, we will not release any clinical data associated with patient identities. 
Instead, we will release our code and provide detailed instructions to replicate our analysis and experiments. 
This study only uses publicly available and de-identified data.

\acks{The authors want to thank the anonymous reviewers for their constructive suggestions. This work was supported by a gift from West Cancer Foundation, Ralph E. Powe Junior Faculty Enhancement Award, and the National Science Foundation with award number IIS-2245920. With gifts from Adobe Research, we purchased a GPU workstation for this study.}

%\bibliography{custom}

\begin{thebibliography}{42}
\providecommand{\natexlab}[1]{#1}
\providecommand{\url}[1]{\texttt{#1}}
\expandafter\ifx\csname urlstyle\endcsname\relax
  \providecommand{\doi}[1]{doi: #1}\else
  \providecommand{\doi}{doi: \begingroup \urlstyle{rm}\Url}\fi

\bibitem[{Bahdanau} et~al.(2016){Bahdanau}, {Brakel}, {Xu}, {Goyal}, {Lowe},
  {Pineau}, {Courville}, and {Bengio}]{bahdanau2016actor}
Dzmitry {Bahdanau}, Philemon {Brakel}, Kelvin {Xu}, Anirudh {Goyal}, Ryan
  {Lowe}, Joelle {Pineau}, Aaron {Courville}, and Yoshua {Bengio}.
\newblock {An Actor-Critic Algorithm for Sequence Prediction}.
\newblock \emph{arXiv e-prints}, art. arXiv:1607.07086, July 2016.
\newblock URL \url{https://arxiv.org/pdf/1607.07086.pdf}.

\bibitem[Chen et~al.(2020)Chen, Song, Chang, and Wan]{chen2020generating}
Zhihong Chen, Yan Song, Tsung-Hui Chang, and Xiang Wan.
\newblock Generating radiology reports via memory-driven transformer.
\newblock In \emph{Proceedings of the 2020 Conference on Empirical Methods in
  Natural Language Processing (EMNLP)}, pages 1439--1449, Online, November
  2020. Association for Computational Linguistics.
\newblock \doi{10.18653/v1/2020.emnlp-main.112}.
\newblock URL \url{https://aclanthology.org/2020.emnlp-main.112}.

\bibitem[Chen et~al.(2021)Chen, Shen, Song, and Wan]{chen2021cross}
Zhihong Chen, Yaling Shen, Yan Song, and Xiang Wan.
\newblock Cross-modal memory networks for radiology report generation.
\newblock In \emph{Proceedings of the 59th Annual Meeting of the Association
  for Computational Linguistics and the 11th International Joint Conference on
  Natural Language Processing (Volume 1: Long Papers)}, pages 5904--5914,
  Online, August 2021. Association for Computational Linguistics.
\newblock \doi{10.18653/v1/2021.acl-long.459}.
\newblock URL \url{https://aclanthology.org/2021.acl-long.459}.

\bibitem[Dalla~Serra et~al.(2022)Dalla~Serra, Clackett, MacKinnon, Wang,
  Deligianni, Dalton, and O{'}Neil]{dalla2022multimodel}
Francesco Dalla~Serra, William Clackett, Hamish MacKinnon, Chaoyang Wang, Fani
  Deligianni, Jeff Dalton, and Alison~Q. O{'}Neil.
\newblock Multimodal generation of radiology reports using knowledge-grounded
  extraction of entities and relations.
\newblock In \emph{Proceedings of the 2nd Conference of the Asia-Pacific
  Chapter of the Association for Computational Linguistics and the 12th
  International Joint Conference on Natural Language Processing (Volume 1: Long
  Papers)}, pages 615--624, Online only, November 2022. Association for
  Computational Linguistics.
\newblock URL \url{https://aclanthology.org/2022.aacl-main.47}.

\bibitem[Delbrouck et~al.(2022)Delbrouck, Chambon, Bluethgen, Tsai, Almusa, and
  Langlotz]{delbrouck2022improving}
Jean-Benoit Delbrouck, Pierre Chambon, Christian Bluethgen, Emily Tsai, Omar
  Almusa, and Curtis Langlotz.
\newblock Improving the factual correctness of radiology report generation with
  semantic rewards.
\newblock In \emph{Findings of the Association for Computational Linguistics:
  EMNLP 2022}, pages 4348--4360, Abu Dhabi, United Arab Emirates, December
  2022. Association for Computational Linguistics.
\newblock URL \url{https://aclanthology.org/2022.findings-emnlp.319}.

\bibitem[Demner-Fushman et~al.(2015)Demner-Fushman, Kohli, Rosenman, Shooshan,
  Rodriguez, Antani, Thoma, and McDonald]{demner2016preparing}
Dina Demner-Fushman, Marc~D. Kohli, Marc~B. Rosenman, Sonya~E. Shooshan,
  Laritza Rodriguez, Sameer Antani, George~R. Thoma, and Clement~J. McDonald.
\newblock Preparing a collection of radiology examinations for distribution and
  retrieval.
\newblock \emph{Journal of the American Medical Informatics Association},
  23\penalty0 (2):\penalty0 304--310, 07 2015.
\newblock ISSN 1067-5027.
\newblock \doi{10.1093/jamia/ocv080}.
\newblock URL \url{https://doi.org/10.1093/jamia/ocv080}.

\bibitem[Denkowski and Lavie(2011)]{denkowski2011meteor}
Michael Denkowski and Alon Lavie.
\newblock Meteor 1.3: Automatic metric for reliable optimization and evaluation
  of machine translation systems.
\newblock In \emph{Proceedings of the Sixth Workshop on Statistical Machine
  Translation}, pages 85--91, Edinburgh, Scotland, July 2011. Association for
  Computational Linguistics.
\newblock URL \url{https://aclanthology.org/W11-2107}.

\bibitem[Hao et~al.(2022)Hao, Liu, and Mou]{hao2022tetacher}
Yongchang Hao, Yuxin Liu, and Lili Mou.
\newblock Teacher forcing recovers reward functions for text generation.
\newblock In Alice~H. Oh, Alekh Agarwal, Danielle Belgrave, and Kyunghyun Cho,
  editors, \emph{Advances in Neural Information Processing Systems}, 2022.
\newblock URL \url{https://openreview.net/forum?id=1_gypPuWUC3}.

\bibitem[Harzig et~al.(2019)Harzig, Chen, Chen, and
  Lienhart]{harzig2019addressing}
Philipp Harzig, Yan-Ying Chen, Francine Chen, and Rainer Lienhart.
\newblock Addressing data bias problems for chest x-ray image report
  generation.
\newblock \emph{arXiv preprint arXiv:1908.02123}, 2019.
\newblock URL \url{https://arxiv.org/pdf/1908.02123.pdf}.

\bibitem[Irvin et~al.(2019)Irvin, Rajpurkar, Ko, Yu, Ciurea-Ilcus, Chute,
  Marklund, Haghgoo, Ball, Shpanskaya, Seekins, Mong, Halabi, Sandberg, Jones,
  Larson, Langlotz, Patel, Lungren, and Ng]{irvin2019chexpert}
Jeremy Irvin, Pranav Rajpurkar, Michael Ko, Yifan Yu, Silviana Ciurea-Ilcus,
  Chris Chute, Henrik Marklund, Behzad Haghgoo, Robyn Ball, Katie Shpanskaya,
  Jayne Seekins, David~A. Mong, Safwan~S. Halabi, Jesse~K. Sandberg, Ricky
  Jones, David~B. Larson, Curtis~P. Langlotz, Bhavik~N. Patel, Matthew~P.
  Lungren, and Andrew~Y. Ng.
\newblock Chexpert: A large chest radiograph dataset with uncertainty labels
  and expert comparison.
\newblock In \emph{Proceedings of the AAAI Conference on Artificial
  Intelligence}, volume~33, pages 590--597, Jul. 2019.
\newblock \doi{10.1609/aaai.v33i01.3301590}.
\newblock URL \url{https://ojs.aaai.org/index.php/AAAI/article/view/3834}.

\bibitem[J~Kurisinkel et~al.(2021)J~Kurisinkel, Aw, and
  Chen]{kurisinkel2021coherent}
Litton J~Kurisinkel, Ai~Ti Aw, and Nancy~F Chen.
\newblock Coherent and concise radiology report generation via context specific
  image representations and orthogonal sentence states.
\newblock In \emph{Proceedings of the 2021 Conference of the North American
  Chapter of the Association for Computational Linguistics: Human Language
  Technologies: Industry Papers}, pages 246--254, Online, June 2021.
  Association for Computational Linguistics.
\newblock \doi{10.18653/v1/2021.naacl-industry.31}.
\newblock URL \url{https://aclanthology.org/2021.naacl-industry.31}.

\bibitem[Jain et~al.(2021)Jain, Agrawal, Saporta, Truong, Duong, Bui, Chambon,
  Zhang, Lungren, Ng, Langlotz, Rajpurkar, and Rajpurkar]{jain2021radgraph}
Saahil Jain, Ashwin Agrawal, Adriel Saporta, Steven Truong, Du~Nguyen Duong,
  Tan Bui, Pierre Chambon, Yuhao Zhang, Matthew Lungren, Andrew Ng, Curtis
  Langlotz, Pranav Rajpurkar, and Pranav Rajpurkar.
\newblock Radgraph: Extracting clinical entities and relations from radiology
  reports.
\newblock In J.~Vanschoren and S.~Yeung, editors, \emph{Proceedings of the
  Neural Information Processing Systems Track on Datasets and Benchmarks},
  volume~1, 2021.
\newblock URL
  \url{https://datasets-benchmarks-proceedings.neurips.cc/paper/2021/file/c8ffe9a587b126f152ed3d89a146b445-Paper-round1.pdf}.

\bibitem[Jing et~al.(2018)Jing, Xie, and Xing]{jing2018automatic}
Baoyu Jing, Pengtao Xie, and Eric Xing.
\newblock On the automatic generation of medical imaging reports.
\newblock In \emph{Proceedings of the 56th Annual Meeting of the Association
  for Computational Linguistics (Volume 1: Long Papers)}, pages 2577--2586,
  Melbourne, Australia, July 2018. Association for Computational Linguistics.
\newblock \doi{10.18653/v1/P18-1240}.
\newblock URL \url{https://aclanthology.org/P18-1240}.

\bibitem[Jing et~al.(2019)Jing, Wang, and Xing]{jing2019show}
Baoyu Jing, Zeya Wang, and Eric Xing.
\newblock Show, describe and conclude: On exploiting the structure information
  of chest {X}-ray reports.
\newblock In \emph{Proceedings of the 57th Annual Meeting of the Association
  for Computational Linguistics}, pages 6570--6580, Florence, Italy, July 2019.
  Association for Computational Linguistics.
\newblock \doi{10.18653/v1/P19-1657}.
\newblock URL \url{https://aclanthology.org/P19-1657}.

\bibitem[Johnson et~al.(2019)Johnson, Pollard, Berkowitz, Greenbaum, Lungren,
  Deng, Mark, and Horng]{johnson2019mimic}
Alistair E~W Johnson, Tom~J Pollard, Seth~J Berkowitz, Nathaniel~R Greenbaum,
  Matthew~P Lungren, Chih-ying Deng, Roger~G Mark, and Steven Horng.
\newblock {MIMIC-CXR, a de-identified publicly available database of chest
  radiographs with free-text reports}.
\newblock \emph{Scientific Data}, 6\penalty0 (1):\penalty0 317, dec 2019.
\newblock ISSN 2052-4463.
\newblock \doi{10.1038/s41597-019-0322-0}.
\newblock URL \url{https://doi.org/10.1038/s41597-019-0322-0}.

\bibitem[Kingma and Ba(2015)]{kingma2014adam}
Diederik~P. Kingma and Jimmy Ba.
\newblock Adam: A method for stochastic optimization.
\newblock In \emph{International Conference on Learning Representations
  (ICLR)}, 2015.
\newblock URL \url{http://arxiv.org/abs/1412.6980}.

\bibitem[Li et~al.(2022)Li, Cai, Verspoor, Pan, Liang, and Chang]{mli2022cross}
M.~Li, W.~Cai, K.~Verspoor, S.~Pan, X.~Liang, and X.~Chang.
\newblock Cross-modal clinical graph transformer for ophthalmic report
  generation.
\newblock In \emph{2022 IEEE/CVF Conference on Computer Vision and Pattern
  Recognition (CVPR)}, pages 20624--20633, Los Alamitos, CA, USA, jun 2022.
  IEEE Computer Society.
\newblock \doi{10.1109/CVPR52688.2022.02000}.
\newblock URL
  \url{https://doi.ieeecomputersociety.org/10.1109/CVPR52688.2022.02000}.

\bibitem[Lin(2004)]{lin2004rouge}
Chin-Yew Lin.
\newblock {ROUGE}: A package for automatic evaluation of summaries.
\newblock In \emph{Text Summarization Branches Out}, pages 74--81, Barcelona,
  Spain, July 2004. Association for Computational Linguistics.
\newblock URL \url{https://aclanthology.org/W04-1013}.

\bibitem[Liu et~al.(2021{\natexlab{a}})Liu, Ge, and Wu]{liu2021competence}
Fenglin Liu, Shen Ge, and Xian Wu.
\newblock Competence-based multimodal curriculum learning for medical report
  generation.
\newblock In \emph{Proceedings of the 59th Annual Meeting of the Association
  for Computational Linguistics and the 11th International Joint Conference on
  Natural Language Processing (Volume 1: Long Papers)}, pages 3001--3012,
  Online, August 2021{\natexlab{a}}. Association for Computational Linguistics.
\newblock \doi{10.18653/v1/2021.acl-long.234}.
\newblock URL \url{https://aclanthology.org/2021.acl-long.234}.

\bibitem[Liu et~al.(2021{\natexlab{b}})Liu, Wu, Ge, Fan, and
  Zou]{liu2021exploring}
Fenglin Liu, Xian Wu, Shen Ge, Wei Fan, and Yuexian Zou.
\newblock Exploring and distilling posterior and prior knowledge for radiology
  report generation.
\newblock In \emph{Proceedings of the IEEE/CVF Conference on Computer Vision
  and Pattern Recognition (CVPR)}, pages 13753--13762, June 2021{\natexlab{b}}.
\newblock URL
  \url{https://openaccess.thecvf.com/content/CVPR2021/html/Liu_Exploring_and_Distilling_Posterior_and_Prior_Knowledge_for_Radiology_Report_CVPR_2021_paper.html}.

\bibitem[Liu et~al.(2021{\natexlab{c}})Liu, Yin, Wu, Ge, Zhang, and
  Sun]{liu2021contrastive}
Fenglin Liu, Changchang Yin, Xian Wu, Shen Ge, Ping Zhang, and Xu~Sun.
\newblock Contrastive attention for automatic chest {X}-ray report generation.
\newblock In \emph{Findings of the Association for Computational Linguistics:
  ACL-IJCNLP 2021}, pages 269--280, Online, August 2021{\natexlab{c}}.
  Association for Computational Linguistics.
\newblock \doi{10.18653/v1/2021.findings-acl.23}.
\newblock URL \url{https://aclanthology.org/2021.findings-acl.23}.

\bibitem[Lovelace and Mortazavi(2020)]{lovelace2020learning}
Justin Lovelace and Bobak Mortazavi.
\newblock Learning to generate clinically coherent chest {X}-ray reports.
\newblock In \emph{Findings of the Association for Computational Linguistics:
  EMNLP 2020}, pages 1235--1243, Online, November 2020. Association for
  Computational Linguistics.
\newblock \doi{10.18653/v1/2020.findings-emnlp.110}.
\newblock URL \url{https://aclanthology.org/2020.findings-emnlp.110}.

\bibitem[Miura et~al.(2021)Miura, Zhang, Tsai, Langlotz, and
  Jurafsky]{miura2021improving}
Yasuhide Miura, Yuhao Zhang, Emily Tsai, Curtis Langlotz, and Dan Jurafsky.
\newblock Improving factual completeness and consistency of image-to-text
  radiology report generation.
\newblock In \emph{Proceedings of the 2021 Conference of the North American
  Chapter of the Association for Computational Linguistics: Human Language
  Technologies}, pages 5288--5304, Online, June 2021. Association for
  Computational Linguistics.
\newblock \doi{10.18653/v1/2021.naacl-main.416}.
\newblock URL \url{https://aclanthology.org/2021.naacl-main.416}.

\bibitem[Mu and Viswanath(2018)]{mu2018allbutthetop}
Jiaqi Mu and Pramod Viswanath.
\newblock All-but-the-top: Simple and effective postprocessing for word
  representations.
\newblock In \emph{International Conference on Learning Representations}, 2018.
\newblock URL \url{https://openreview.net/forum?id=HkuGJ3kCb}.

\bibitem[Nguyen et~al.(2021)Nguyen, Nie, Badamdorj, Liu, Zhu, Truong, and
  Cheng]{nguyen2021automated}
Hoang Nguyen, Dong Nie, Taivanbat Badamdorj, Yujie Liu, Yingying Zhu, Jason
  Truong, and Li~Cheng.
\newblock Automated generation of accurate {\&} fluent medical {X}-ray reports.
\newblock In \emph{Proceedings of the 2021 Conference on Empirical Methods in
  Natural Language Processing}, pages 3552--3569, Online and Punta Cana,
  Dominican Republic, November 2021. Association for Computational Linguistics.
\newblock \doi{10.18653/v1/2021.emnlp-main.288}.
\newblock URL \url{https://aclanthology.org/2021.emnlp-main.288}.

\bibitem[Nikkarinen et~al.(2021)Nikkarinen, Pimentel, Blasi, and
  Cotterell]{nikkarinen2021modeling}
Irene Nikkarinen, Tiago Pimentel, Dami{\'a}n Blasi, and Ryan Cotterell.
\newblock Modeling the unigram distribution.
\newblock In \emph{Findings of the Association for Computational Linguistics:
  ACL-IJCNLP 2021}, pages 3721--3729, Online, August 2021. Association for
  Computational Linguistics.
\newblock \doi{10.18653/v1/2021.findings-acl.326}.
\newblock URL \url{https://aclanthology.org/2021.findings-acl.326}.

\bibitem[Nishino et~al.(2020)Nishino, Ozaki, Momoki, Taniguchi, Kano, Nakano,
  Tagawa, Taniguchi, Ohkuma, and Nakamura]{nishino2020reinforcement}
Toru Nishino, Ryota Ozaki, Yohei Momoki, Tomoki Taniguchi, Ryuji Kano, Norihisa
  Nakano, Yuki Tagawa, Motoki Taniguchi, Tomoko Ohkuma, and Keigo Nakamura.
\newblock Reinforcement learning with imbalanced dataset for data-to-text
  medical report generation.
\newblock In \emph{Findings of the Association for Computational Linguistics:
  EMNLP 2020}, pages 2223--2236, Online, November 2020. Association for
  Computational Linguistics.
\newblock \doi{10.18653/v1/2020.findings-emnlp.202}.
\newblock URL \url{https://aclanthology.org/2020.findings-emnlp.202}.

\bibitem[Papineni et~al.(2002)Papineni, Roukos, Ward, and
  Zhu]{papineni2002bleu}
Kishore Papineni, Salim Roukos, Todd Ward, and Wei-Jing Zhu.
\newblock Bleu: A method for automatic evaluation of machine translation.
\newblock In \emph{Proceedings of the 40th Annual Meeting on Association for
  Computational Linguistics}, ACL '02, page 311–318, USA, 2002. Association
  for Computational Linguistics.
\newblock \doi{10.3115/1073083.1073135}.
\newblock URL \url{https://doi.org/10.3115/1073083.1073135}.

\bibitem[Paszke et~al.(2019)Paszke, Gross, Massa, Lerer, Bradbury, Chanan,
  Killeen, Lin, Gimelshein, Antiga, Desmaison, Kopf, Yang, DeVito, Raison,
  Tejani, Chilamkurthy, Steiner, Fang, Bai, and Chintala]{adam2019pytorch}
Adam Paszke, Sam Gross, Francisco Massa, Adam Lerer, James Bradbury, Gregory
  Chanan, Trevor Killeen, Zeming Lin, Natalia Gimelshein, Luca Antiga, Alban
  Desmaison, Andreas Kopf, Edward Yang, Zachary DeVito, Martin Raison, Alykhan
  Tejani, Sasank Chilamkurthy, Benoit Steiner, Lu~Fang, Junjie Bai, and Soumith
  Chintala.
\newblock Pytorch: An imperative style, high-performance deep learning library.
\newblock In \emph{Advances in Neural Information Processing Systems 32},
  volume~32, pages 8024--8035. Curran Associates, 2019.
\newblock URL
  \url{https://proceedings.neurips.cc/paper/2019/file/bdbca288fee7f92f2bfa9f7012727740-Paper.pdf}.

\bibitem[Qin and Song(2022)]{qin2022reinforced}
Han Qin and Yan Song.
\newblock Reinforced cross-modal alignment for radiology report generation.
\newblock In \emph{Findings of the Association for Computational Linguistics:
  ACL 2022}, pages 448--458, Dublin, Ireland, May 2022. Association for
  Computational Linguistics.
\newblock \doi{10.18653/v1/2022.findings-acl.38}.
\newblock URL \url{https://aclanthology.org/2022.findings-acl.38}.

\bibitem[Shi et~al.(2018)Shi, Chen, Qiu, and Huang]{shi2018toward}
Zhan Shi, Xinchi Chen, Xipeng Qiu, and Xuanjing Huang.
\newblock Toward diverse text generation with inverse reinforcement learning.
\newblock In \emph{Proceedings of the 27th International Joint Conference on
  Artificial Intelligence}, IJCAI'18, page 4361–4367. AAAI Press, 2018.
\newblock ISBN 9780999241127.
\newblock URL \url{https://arxiv.org/abs/1804.11258}.

\bibitem[Smit et~al.(2020)Smit, Jain, Rajpurkar, Pareek, Ng, and
  Lungren]{smit2020combining}
Akshay Smit, Saahil Jain, Pranav Rajpurkar, Anuj Pareek, Andrew Ng, and Matthew
  Lungren.
\newblock Combining automatic labelers and expert annotations for accurate
  radiology report labeling using {BERT}.
\newblock In \emph{Proceedings of the 2020 Conference on Empirical Methods in
  Natural Language Processing (EMNLP)}, pages 1500--1519, Online, November
  2020. Association for Computational Linguistics.
\newblock \doi{10.18653/v1/2020.emnlp-main.117}.
\newblock URL \url{https://aclanthology.org/2020.emnlp-main.117}.

\bibitem[Sutskever et~al.(2014)Sutskever, Vinyals, and
  Le]{sutskever2014advances}
Ilya Sutskever, Oriol Vinyals, and Quoc~V Le.
\newblock Sequence to sequence learning with neural networks.
\newblock In Z.~Ghahramani, M.~Welling, C.~Cortes, N.~Lawrence, and K.Q.
  Weinberger, editors, \emph{Advances in Neural Information Processing
  Systems}, volume~27. Curran Associates, Inc., 2014.
\newblock URL
  \url{https://proceedings.neurips.cc/paper/2014/file/a14ac55a4f27472c5d894ec1c3c743d2-Paper.pdf}.

\bibitem[Sutton et~al.(1999)Sutton, McAllester, Singh, and
  Mansour]{sutton1999policy}
Richard~S. Sutton, David McAllester, Satinder Singh, and Yishay Mansour.
\newblock Policy gradient methods for reinforcement learning with function
  approximation.
\newblock In \emph{Advances in Neural Information Processing Systems}, NIPS'99,
  page 1057–1063, Cambridge, MA, USA, 1999. MIT Press.
\newblock URL
  \url{https://proceedings.neurips.cc/paper/1999/file/464d828b85b0bed98e80ade0a5c43b0f-Paper.pdf}.

\bibitem[Tian et~al.(2021)Tian, Chen, Zhang, Feng, Xiong, Wu, and
  Dou]{tian2021embedding}
Jiachen Tian, Shizhan Chen, Xiaowang Zhang, Zhiyong Feng, Deyi Xiong, Shaojuan
  Wu, and Chunliu Dou.
\newblock Re-embedding difficult samples via mutual information constrained
  semantically oversampling for imbalanced text classification.
\newblock In \emph{Proceedings of the 2021 Conference on Empirical Methods in
  Natural Language Processing}, pages 3148--3161, Online and Punta Cana,
  Dominican Republic, November 2021. Association for Computational Linguistics.
\newblock \doi{10.18653/v1/2021.emnlp-main.252}.
\newblock URL \url{https://aclanthology.org/2021.emnlp-main.252}.

\bibitem[Vaswani et~al.(2017)Vaswani, Shazeer, Parmar, Uszkoreit, Jones, Gomez,
  Kaiser, and Polosukhin]{vaswani2017attention}
Ashish Vaswani, Noam Shazeer, Niki Parmar, Jakob Uszkoreit, Llion Jones,
  Aidan~N. Gomez, \L{}ukasz Kaiser, and Illia Polosukhin.
\newblock Attention is all you need.
\newblock In \emph{Proceedings of the 31st International Conference on Neural
  Information Processing Systems}, NIPS'17, page 6000–6010, Red Hook, NY,
  USA, 2017. Curran Associates Inc.
\newblock ISBN 9781510860964.
\newblock URL
  \url{https://proceedings.neurips.cc/paper/2017/file/3f5ee243547dee91fbd053c1c4a845aa-Paper.pdf}.

\bibitem[Wang et~al.(2020)Wang, Huang, Huang, Hu, Wang, and
  Gu]{wang2020Improving}
Lingxiao Wang, Jing Huang, Kevin Huang, Ziniu Hu, Guangtao Wang, and Quanquan
  Gu.
\newblock Improving neural language generation with spectrum control.
\newblock In \emph{International Conference on Learning Representations}, 2020.
\newblock URL \url{https://openreview.net/forum?id=ByxY8CNtvr}.

\bibitem[Welleck et~al.(2020)Welleck, Kulikov, Roller, Dinan, Cho, and
  Weston]{sean2019neural}
Sean Welleck, Ilia Kulikov, Stephen Roller, Emily Dinan, Kyunghyun Cho, and
  Jason Weston.
\newblock Neural text generation with unlikelihood training.
\newblock In \emph{ICLR}, 2020.
\newblock URL \url{https://openreview.net/forum?id=SJeYe0NtvH}.

\bibitem[Wu and Huang(2022)]{wu2022unsupervised}
Yuexin Wu and Xiaolei Huang.
\newblock Unsupervised reinforcement adaptation for class-imbalanced text
  classification.
\newblock In \emph{Proceedings of the 11th Joint Conference on Lexical and
  Computational Semantics}, pages 311--322, Seattle, Washington, July 2022.
  Association for Computational Linguistics.
\newblock \doi{10.18653/v1/2022.starsem-1.27}.
\newblock URL \url{https://aclanthology.org/2022.starsem-1.27}.

\bibitem[Yang et~al.(2022)Yang, Wu, Ge, Zhou, and Xiao]{yang2022knowledge}
Shuxin Yang, Xian Wu, Shen Ge, S~Kevin Zhou, and Li~Xiao.
\newblock Knowledge matters: Chest radiology report generation with general and
  specific knowledge.
\newblock \emph{Medical image analysis}, 80:\penalty0 102510, August 2022.
\newblock ISSN 1361-8415.
\newblock \doi{10.1016/j.media.2022.102510}.
\newblock URL \url{https://doi.org/10.1016/j.media.2022.102510}.

\bibitem[Yang et~al.(2020)Yang, Li, Fukumoto, and Ye]{yang2020hscnn}
Wenshuo Yang, Jiyi Li, Fumiyo Fukumoto, and Yanming Ye.
\newblock {HSCNN}: A hybrid-{S}iamese convolutional neural network for
  extremely imbalanced multi-label text classification.
\newblock In \emph{Proceedings of the 2020 Conference on Empirical Methods in
  Natural Language Processing (EMNLP)}, pages 6716--6722, Online, November
  2020. Association for Computational Linguistics.
\newblock \doi{10.18653/v1/2020.emnlp-main.545}.
\newblock URL \url{https://aclanthology.org/2020.emnlp-main.545}.

\bibitem[Yu et~al.(2022)Yu, Song, Kim, Lee, Ryu, and Yoon]{yu2022rare}
Sangwon Yu, Jongyoon Song, Heeseung Kim, Seongmin Lee, Woo-Jong Ryu, and
  Sungroh Yoon.
\newblock Rare tokens degenerate all tokens: Improving neural text generation
  via adaptive gradient gating for rare token embeddings.
\newblock In \emph{Proceedings of the 60th Annual Meeting of the Association
  for Computational Linguistics (Volume 1: Long Papers)}, pages 29--45, Dublin,
  Ireland, May 2022. Association for Computational Linguistics.
\newblock \doi{10.18653/v1/2022.acl-long.3}.
\newblock URL \url{https://aclanthology.org/2022.acl-long.3}.

\end{thebibliography}

\appendix

\begin{table*}[htp]
\centering 
\caption{Summary of TIMER's performance improvements over baselines. $\hat{\Delta}$ indicates percentage improvements over baselines.} \label{tab:delta_all}
\resizebox{.78\textwidth}{!}{
    \begin{tabular}{c||ccccccc}
        Methods $(\%)$ & BLEU\_1 &BLEU\_2 & BLEU\_3 & BLEU\_4 & Meteor& Rouge\_L & Clinical Metric \\\hline\hline
        \multicolumn{7}{c}{IU X-RAY} \\ \hline
        $\hat{\Delta}$ &6.42	&7.57	&9.07	&13.06	&4.45	&4.55	&61.16 \\
        $\hat{\Delta}$-BiLSTM  &17.96	&10.89	&12.10	&20.14	&8.72	&11.64  &45.09\\
        $\hat{\Delta}$-R2GEN &1.11 &1.75	&2.58	&5.02	&0.84 &3.10&48.38 \\
        $\hat{\Delta}$-CMN &8.37	&10.13	&11.04	&12.58	&7.32	&4.00 &45.61 \\
        $\hat{\Delta}$-CMM+RL &0.08   &8.01	&11.14	&15.59	&1.39	&0.13 &131.42\\
        \hline\hline
        \multicolumn{7}{c}{MIMIC} \\ \hline
        $\hat{\Delta}$ &11.27	&9.66	&9.01	&10.50	&8.64	&3.54	&49.30   \\
        $\hat{\Delta}$-BiLSTM &42.86	&42.61	&44.27	&48.57	&30.55 &7.69 &53.25 \\
        $\hat{\Delta}$-R2GEN &8.13 &2.27	&0.69	&0.97	&6.91	&2.79 &118.17 \\
        $\hat{\Delta}$-CMN &7.58	&5.04	&3.77	&4.94	&3.67	&3.17 &84.08\\
        $\hat{\Delta}$-CMM+RL &0.52	&1.76 &1.03 &3.79 &1.17 &1.23 &167.48\\
    \end{tabular}
}
\end{table*}

\section{Preprocessing Details}
\label{sec:prepro}
Following the preprocessing of R2Gen,  the raw documents are converted to lowercase and tokenized using the NLTK library. Furthermore, we removed redundant spaces, empty lines, serial numbers, and punctuation marks from the documents 
In IU X-RAY, we apply the widely-used splits in~\citep{chen2020generating, jing2019show, liu2021exploring}  and partition the dataset into train/validation/test set by 7:1:2. Following the works \citep{chen2020generating, chen2021cross}, we remove the tokens whose frequency of occurrence in the training set is less than 3, resulting in 789 words for the entire dataset.
In MIMIC, we adopt the official splits for the  dataset to report our results: 368,960 for training, 2,991 for validation, and 5,159 for test. 
Following the works~\citep{chen2020generating, chen2021cross, liu2021contrastive}, we keep the tokens with a frequency in the training set are more than 10.

\section{Implementation Details}
\label{sec:impl}

% We adopt CMN as our text generation model and evaluate our algorithm on this model. The model parameter and learning rate follow the paper's setting. 
We use ADAM~\citep{kingma2014adam} optimizer to train our model with the learning rate $0.001$ and decay such rate by a factor of $0.9$ per epoch for each dataset. 
We update the TIMER for each inner loop training in the IU X-RAY dataset and every 100 iterations of inner loop training in the MIMIC dataset. 
The max training epoch is 100 for the IU X-RAY and 30 for the MIMIC, due to the data sizes and our computational resources.
We generate tokens by beam search~\citep{sutskever2014advances} with 3 beam size in the test stage for all experiments.
All implementations are on PyTorch~\citep{adam2019pytorch}. 

In baseline BiLSTM, we set the number of tags for semantic attention as 10 and all hidden states and word embeddings as 512. The learning
rates for the CNN and the hierarchical LSTM are 1e-5 and 5e-4 respectively.

In baseline R2GEN,  We adopt the ResNet101 to extract images' features with the dimension of each feature set to 2048. The dimension of relational memory in multi-head attention is  512 and contains 8 heads. The number of memory slots is set to 3 by default. 
The learning rate is 5e-5 for the visual extractor and 
1e-4 for other parameters. We decay such rate by a
factor of 0.8 per epoch for each dataset.

In baseline CMN, the image feature extractor has the same setting as R2GE. The encoder-decoder structure adopts Transformer with 3 layers and 8 attention heads.
The memory matrix dimension is 512 and  the number of memory vectors is set to 2048. 
The learning rates of the visual extractor and other
parameters are set to $5 \times 10^{-5}$ and $10^{-4}$, respectively, and we decay them by a 0.8 rate per epoch for all datasets.

CMM+RL baseline keeps all the same settings as CMN. Following the setting in the paper, we adopt the greedy sampling method to generate reports for self-critical learning. We set the batch size as 8 since this achieves the best result in the paper's report.

\end{document}